# Complementary Advantages of ChatGPTs and Human Readers in Reasoning: Evidence from English Text Reading Comprehension

**Tongquan Zhou**[1*]
Southeast University
**Siyi Cao**[1]
Southeast University
**Tao Wang**[3,**]
Qufu Normal University

**Yao Zhang**[1*,**]
Southeast University
**Yulu Li**[2]
Qufu Normal University

## Abstract

ChatGPT has shown its great power in text processing, including its reasoning ability from text reading. However, there has not been any direct comparison between human readers and ChatGPT in reasoning ability related to text reading. This study was undertaken to investigate how ChatGPTs (i.e., ChatGPT and ChatGPT Plus) and Chinese senior school students as ESL learners exhibited their reasoning ability from English narrative texts. Additionally, we compared the two ChatGPTs in the reasoning performances when commands were updated elaborately.

The whole study was composed of three reasoning tests: Test 1 for commonsense inference, Test 2 for emotional inference, and Test 3 for causal inference. The results showed that in Test 1, the students outdid the two ChatGPT versions in local-culture-related inferences but performed worse than the chatbots in daily-life inferences. In Test 2, ChatGPT Plus excelled whereas ChatGPT lagged behind in accuracy. In association with both accuracy and frequency of correct responses, the students were inferior to the two chatbots. Compared with ChatGPTs' better performance in positive emotions, the students showed their superiority in inferring negative emotions. In Test 3, the students demonstrated better logical analysis, outdoing both chatbots. In updating command condition, ChatGPT Plus displayed good causal reasoning ability while ChatGPT kept unchanged.

Our study reveals that human readers and ChatGPTs have their respective advantages and disadvantages in drawing inferences from text reading comprehension, unlocking a complementary relationship in text-based reasoning.

## 1. Introduction

Reading comprehension is an ability to understand text's meaning or learn from a provided text, which works in connection with various skills, including automaticity, higher-level language comprehension processes, background knowledge, schema construction, knowledge of text structures, the capacity of different memory structures, and inference-making (Basaraba et al., 2013; Toprak & Cakir, 2021). A good reading comprehension is more often than not pertinent to a reader's correct inference making. Inference-making refers to the capacity to interpret implicit information in the text (Martinez-Lincoln et al., 2021), acting as a crucial component of reading comprehension capacity (Clinton et al., 2020; Martinez-Lincoln et al., 2021).

[1] School of Foreign Languages, Southeast University, Nanjing, China, 211189
[2] College of Chinese Literature and Language, Qufu Normal University, Qufu, China, 273165
[3] School of Psychology, Qufu Normal University, Qufu, China, 273165
* First authorship
** Corresponding authors

Human readers comprehend a text by integrating complicated linguistic and cognitive processes. They need to recognize words and sentences before connecting them to construct the underlying meaning and coherent representation of the text (Kendeou et al., 2014). This process requires not only cognitive architecture and cognitive procedures, such as working memory and retrieval operations but also prior knowledge, such as word knowledge (Perfetti & Stafura, 2014). During this process, inferential comprehension skills play an important role for readers to reach proficient second language reading comprehension (Perkins, 1988). Previous research has confirmed the direct and significant influence of inference-making capacity on reading comprehension (Ahmed et al., 2016; Oslund et al., 2016). The inference-making ability is related to comprehension skills (Ahmed et al., 2016; Barnes et al., 2015). More specifically, inferential ability could predict performance in reading comprehension (Ahmed et al., 2016) and reversely, inference-making ability is influenced by reading comprehension ability (Barnes et al., 2015; Li & Kirby, 2014). Other factors that have an impact on inferential ability include vocabulary knowledge (Oslund et al., 2016; Prior et al., 2014), L2 word reading skills and higher cognitive processes (Prior et al., 2014), inference instructions (Hall et al., 2020), and teachers' knowledge of reading and teaching skills (Westbrook et al., 2019).

The inference/reasoning ability of LLMs originates from the chain of thought prompting (CoT) and self-consistency strategy (X. Wang et al., 2023; Wei et al., 2023). CoT enables LLMs to infer examples instead of "standard question and answer examples", divide the complicated reasoning process into various easier steps (Kojima et al., 2023, p. 2), and reduce repetitiveness in the coding process and stochasticity of answer generation (X. Wang et al., 2023). Specifically, the CoT greatly improves LLMs' reasoning ability and makes LLMs plausible to deal with math problems, commonsense reasoning, and symbolic manipulation with higher accuracy (X. Wang et al., 2023; Wei et al., 2023). Different from traditional CoT which adopts one decoding path to process tasks, self-consistency deals with prompting by generating multiple reasoning paths to aggregate a final and best answer (X. Wang et al., 2023). This method derives from human experience "if multiple different ways of thinking lead to the same answer, one has greater confidence that the final answer is correct" (X. Wang et al., 2023, p. 1). Accordingly, equipped with the two complementary mechanisms, the LLMs particularly ChatGPT and its updated version could deal with reasoning tasks more accurately.

A set of criteria have been adopted to categorize inferences. For example, Van Den Broek et al. (1993) classified inferences in terms of their functions in maintaining coherence and organizing sources of information. The inferences in his study contained four types: backward inferences, forward elaborations, orthogonal elaborations, and associative inferences. Singer and Ferreira (1983) divided inferences into forward and backward inferences according to the direction either connecting prior text or predicting subsequent plot. Furthermore, inferences were also categorized into inductive, deductive, and analogical inferences on the basis of logical form (Kintsch, 1993). Apart from these, researchers also identified text-based and knowledge-based inferences from the perspective of source of text information or background knowledge (Basaraba et al., 2013; Clinton et al., 2020), the classification to be adopted in this study.

Graesser and colleagues have proposed the unique classification of inferences to analyze narrative texts. According to Magliano and Graesser (1991), there are eleven classes of inferences in light of inference generation during narrative comprehension. Afterward, Graesser et al. (1994) added two more classes (class 12 and 13) to result in a full classification of 13 types of inferences, as shown in Table 1. Among the 13 inferences, causal inference (including causal antecedent and causal consequence), the author's intent or attitude, and the character's emotional reaction were proven to be the most frequently analyzed aspects in the current studies. Meanwhile, the commonsense inference though not covered in the 13 inferences is another frequent type, involving comprehending and deducing the world knowledge that we have to judge and predict new situation so as to make new conclusion (Bang et al., 2023; Storks, 2019). Accordingly, the present study focused on the three text-based inferences (commonsense inference, emotional inference, and causal inference) to investigate how the senior students, ChatGPT, and ChatGPT Plus exhibit their reasoning ability in practice.

| Classes | Type of inference | Brief description |
| --- | --- | --- |
| 1 | Referential | A word or phrase is referentially tied to a previous element or constituent in the text (explicit or inferred). |
| 2 | Causal antecedent | The inference is on a causal chain (bridge) between the current explicit action, event, or state and the previous passage context. |
| 3 | Causal consequence | The inference is on a forecasted causal chain, including physical events and new plans of agents. |
| 4 | Instrument | The inference is an object, part of the body, or resource used when an agent executes an intentional action. |
| 5 | Instantiation of Noun category | The inference is a subcategory or a particular exemplar that instantiates an explicit noun. |
| 6 | Superordinate goal | The inference is a goal that motivates an agent's intentional action. |
| 7 | Subordinate goal/action | The inference is a goal, plan, or action that specifies how an agent's action is achieved. |
| 8 | State | The inference is an ongoing state, from the time frame of the text. The states include an agent's traits, knowledge, and beliefs; the properties of objects and concepts; and the spatial location of entities. |
| 9 | Thematic | This is a main point or moral of the text. |
| 10 | Emotion of reader | The inference is the emotion that the reader experiences when reading a text. |
| 11 | Author's intent or attitude | The inference is the author's attitude or motive in writing a text segment |
| 12 | Case structure role assignment | An explicit noun phrase is assigned to a particular case structure role, e.g., agent, recipient, object, location, time. |
| 13 | Character emotional reaction | The inference is an emotion experienced by a character, caused by or in response to an event or action |

Table 1. Thirteen Classes of Inferences (adapted from Graesser et al. 1994)

Commonsense knowledge is important when readers need to activate implicit inferences such as cause, antecedents, and emotion detection and understand the narrative (Ghosal et al., 2022; Rashkin et al., 2019). Therefore, previous research leveraged abundant resources to empower LLMs with the ability to make commonsense inferences (e.g., Lin et al. 2019, Rashkin et al. 2019, Ghosal et al. 2022) and better LLMs' language process capacity (e.g., Zhao et al. 2023). Thanks to large commonsense knowledge datasets, pre-trained language models have been testified to possess commonsense inference ability (P. Wang et al., 2021).

Emotional inferences in LLMs contain sentiment analysis relating to readers' or authors' attitudes to the plot and emotion detection focusing on the characters' physical reactions (Mao et al., 2022). They are one of knowledge-based inferences that requires readers to activate personal experiences and world knowledge to incorporate such knowledge into establishing the implicit text meaning (Basaraba et al., 2013; Clinton et al., 2020; Graesser et al., 1994; Graesser & Kreuz, 1993). A couple of studies have been conducted on investigating how ChatGPTs infer emotions or sentiments in the comparative perspective (e.g., Bang et al. 2023, Guo et al. 2023, Qin et al. 2023). However, most emotion studies concentrate on the positive, negative, and neutral emotions (i.e., sentiment analysis) in single sentences but not elaborate on the emotion categories in detailed and complex contexts (Yang et al., 2023).

Causal inferences, including causal antecedent and consequence, provide connections between previously-obtained information and the given reading text to construct local coherence within the text (Basaraba et al., 2013; Clinton et al., 2020). The inferences require strict sequential logic, that is, the antecedent never occurs after its consequence and never disappears until the consequence happens (Van Den Broek, 1990). Furthermore, to fully understand a text, one needs to activate causal inferences to connect all successive events, no matter implicit or explicit, to create a coherent plot (Mason & Just, 2004). Previous studies have exerted efforts in developing causal inference models for machine learning (e.g., Pearl 2010, Egami et al. 2018, Yao et al. 2021) and applied the models to different areas under the construction of machine learning (e.g., Hair and Sarstedt 2021, Siebert 2023). These indicate that LLMs have owned certain causal inference abilities.

Human and ChatGPT resort to different mechanisms to make inferences. For human readers, they infer implicit information through complex cognitive processes, such as "synthesizing, generalizing, summarizing, and extrapolating" (Saadatnia et al., 2017, p. 1091), then relate the content to their reasoning and logical extension competence, and prior knowledge (Basaraba et al., 2013; L. Lin et al., 2021; Saadatnia et al., 2017). Besides, the readers engaged in inferential comprehension are required to "recognize and understand the relationships that exist among objects, events, or characters in the text" and draw conclusions by analyzing the structure of texts (Alonzo et al., 2009, p. 35). That is, good readers should be equipped with inferential skills in retrieving background knowledge to achieve text coherence and to fill in missing information that may affect comprehension (Clinton & Van Den Broek, 2012). By contrast, ChatGPT (built on GPT-3.5), a sophisticated chatbot based on large language models (LLMs), is developed to "understand and interpret user requests and then generate appropriate responses in nearly natural human language" (Lund & Wang, 2023). LLMs are well-trained deep-learning models based on a wide range of online texts from different sources, including Wikipedia, news, books, websites, and social media (Ray, 2023; Zhou et al., 2023). These datasets enable LLMs involving ChatGPT to learn the patterns and relationships existing in language, consequently creating responses to a diversity of language-related tasks, such as text analysis, translation, and writing (Ray, 2023). Furthermore, when making inferences, LLMs tend to make conclusions by extracting some trigger words. For example, they infer negative attitudes hidden in the text through identifying trigger words like "but", and "sorry" (Guo et al., 2023).

Currently, the latest version ChatGPT Plus (built on GPT-4) has achieved a great advance in many aspects. According to OpenAI (2023), GPT-4 and GPT 3.5 scored 710 and 670 (out of 800) in SAT (Scholastic Assessment Test) Evidence-based Reading & Writing respectively, suggesting the two versions are capable of processing difficult reading tasks with great accuracy. Besides, both versions performed well on many other tests, for instance, to have a verbal-linguistic IQ of higher than 147 (OpenAI, 2023; Ray, 2023), demonstrating their excellent proficiency in comprehending reading materials. The most distinguishable feature of ChatGPT Plus lies in its broader world knowledge, better problem-solving capacity, and greater reasoning ability (OpenAI, 2023). On these bases, the updated chatbot has achieved remarkably higher scores in many exams that are designed for humans, such as SAT, than GPT-3.5 (OpenAI, 2023). Namely, GPT-4 outperforms GPT-3.5 in analyzing complex texts and making inferences.

The reasoning/inference-making ability of ChatGPT and ChatGPT Plus has received wide attention (e.g, Bang et al. 2023, Liu et al. 2023, Qin et al. 2023). Liu et al. (2023) evaluated the logical reasoning ability of ChatGPT and ChatGPT Plus on various logical reasoning datasets, with the results revealing the two ChatGPTs' impressive logical reasoning ability. Similarly, in Qin et al.'s (2023) reasoning ability test (including arithmetic, commonsense, symbolic, logical reasoning, natural language inference, sentiment analysis, summarization ability, named entity recognition, and dialogic ability), ChatGPT did not always predict correct answers in commonsense reasoning assignments but achieved high scores in entailing premises and hypotheses, demonstrating its good capability in inferring sentence relations and coherence. Bang et al. (2023) explored ChatGPT's multitasking, multilingual, and multi-modal to discuss its strengths and limitations and found that ChatGPT was more skillful at drawing specific conclusions from the general premises but showing weakness in figuring out the rules in the given information and making correct conclusions. Furthermore, Zhu et al. (2023) also suggested that ChatGPT is good at processing objective cases instead of subjective cases. Additionally, ChatGPT displayed good performance at commonsense reasoning concerning the daily experience.

In addition, three recent studies disclosed the comparison between ChatGPT and human in inferring emotions. Elyoseph et al. (2023) adopted performance-based test to investigate ChatGPT's ability in identifying and describing emotions and compared their performances with human emotional data collected by Nandrino et al. (2013). Their findings showed that ChatGPT outdid general population on evaluating emotions and ChatGPT's ability would be improved over time. However, the scale in their study tested mainly four emotions: anger, fear, happiness, or sadness (Nandrino et al., 2013). Since human beings are sentimental creatures, more emotions needed to be testified for a thorough comparison obviously. Kocon et al. (2023, page 9) compared the performances of emotional recognition between human and ChatGPT, indicating that ChatGPT was "Jack of all trades, master of none". This was because ChatGPT may be good at parts of their tests but not excelled human in all datasets, revealing ChatGPT is unstable in inferring emotions.

ESL (English as a second language) learners' inference ability has been explored widely. For example, Gillioz et al. (2012) tested the relationship between individual differences and emotional inferences, finding that individual differences did influence ESL students' inferring results. According to Norouzi et al. (2013), only inferential questions among the tested types influenced the reading comprehension by all the EFL learners with low, immediate, and high proficiency levels. In Jang's (2009) study, diagnostic inferences were investigated from the perspective of cognitive skills, revealing that background information were important in making inferences. Among the other studies are mainly those concerned with lexical inferences, such as the relationship between L2 vocabulary knowledge and lexical inferencing strategy use (e.g., Nassaji 2003, 2004, Parel 2004), the relationship between lexical inference and word structures and context (e.g., Zhang and Koda 2012, Hamada 2014), correlation of L2 word inference success with strategy use (Hamada, 2009), and influence of reading proficiency on lexical inference (Kaivanpanah & Soltani Moghaddam, 2012). Although these studies did not directly announce students' inferential level, their results implied that vocabulary and lexical inferences were significant in the reading process and EFL learners did have the ability to draw inferences from second-language texts.

Different from the previous researches, the present study invited the students of grade two in a senior high school in China to test their commonsense, emotional, and causal inferences. By the time the students took our tests, they had learned English for at least 6 years, studied at least 4500 words, and intensively read 62 long texts. Under the teachers' everyday teaching and guidance, they practiced how to do text-based reading comprehension and knew the basic strategies and skills (including inferential skills) relating to reading comprehension. So to speak, the students approximated immediate English proficiency level. Furthermore, the students who have lived in nearby cities should have heard or experienced the cultures in the commonsense test. Against this background, an interesting issue is whether these students do better than ChatGPT in terms of reading inferential ability.

To summarize, the two ChatGPTs are adept at making inferences. By comparison, although the Chinese students as ESL human readers in the present study did not possess a large vocabulary and text-based knowledge, they should have acquired inferential ability after years' of English study to a large extent. But the research gaps are obvious as below.

Firstly, ChatGPTs performed unsteadily in commonsense inferences and emotional inferences, for the required knowledge is supposed not to be within the chatbots' training data (Mahowald et al., 2023), such as the local customs or cultures in a specific area. Secondly, it remains unknown whether and how ChatGPT and humans can exhibit similar reasoning abilities by inferring the emotions of characters involved in given stories. Thirdly, ChatGPT Plus is reported to be more powerful than ChatGPT but more evidence is required to demonstrate the claim.

Backgrounded by the above, the present study was conducted to examine how ChatGPTs (i.e., ChatGPT and its updated versions) and Chinese high school students as ESL learners exhibit their reasoning ability from English narrative texts. In addition, we compared ChatGPT with ChatGPT Plus (i.e., the updated version) in the reasoning performances by updating commands. Specifically, we attempted to answer the following three questions:

(1) How did the two ChatGPTs and Chinese high school students show their respective inference capacity when involved in narrative text reading?

(2) What advantages and disadvantages could both ChatGPTs and the students show in drawing the three inferences (commonsense inferences, emotional inferences and causal inferences) from English reading comprehension?

(3) What inference changes (if any) might occur in ChatGPT and ChatGPT Plus when elaborate commands were updated?

## 2. Method

### 2.1 Participants

114 senior-2 students (38 females and 76 males) in a middle-level high school (English scores ranked about 12-14 in the citywide unified examinations) in China, voluntarily participated in this study and their ages are around 17. These students have learnt English for at least 6 years and are currently learning English for Gaokao (the College Entrance Examination in China). Before this survey, they had learnt six out of the seven required textbooks, containing 186 texts that had been intensively read and roughly 2300 new words, suggesting they should have mastered nearly 4150 words (about 1900 words had been learned in the high school stage). On this account, we believed that the students had reached a low-intermediate level of English proficiency.

## 2.2 Materials

### 2.2.1 Test 1 Commonsense inference test

Test 1 was designed to test whether the participants and the two versions of ChatGPT would judge the correctness of the sentence containing specific commonsense concerning local culture and discover whether they could judge the daily life from the given context. Therefore, we designed a test containing two parts containing 28 questions, with 13 about local characteristic cultures and 15 concerning daily life experiences. The first type contains local characteristic cultures, including famous persons in China and local customs in Fuzhou, a capital city with various special customs, in the southeast part of China, and the second sort consists of the commonsense relevant to the daily life topics chosen from multiple choices in different Gaokao[4] mocks from different regions of China. Prior to examining the students' commonsense inference, we assessed the test's split-half reliability, which has been verified to be reliable with small language samples (Cole et al., 1989). The statistics showed that the split-half reliability (Spearman-Brown coefficient) of the first part and the second part was 0.710 and 0.701 respectively, revealing that the materials for test were valid. Table 2 illustrates the examples for commonsense test in this study.

---

I. Please judge whether the following statements are True or False, and state your reasons briefly

Q1. As reported by Xinhua News Agency, Yuan Longping is going to deliver a speech at our school on June 13$^{th}$, 2023.
   (Key: F. Yuan Longping passed away in 2021, so he cannot deliver a speech in 2023.)

Q6. On the night of the Mid-Autumn festival, all families in Fuzhou prepare dice and bowls to play the mooncake gambling.
   (Key: F. The mooncake gambling is a custom popular in Xiamen, not in Fuzhou. Therefore, it would not be possible for all families in Fuzhou to celebrate this custom.)

II. Please choose the best answer from A, B, C, or D that is most suitable for the context.
Q14. Salina Joe began to ____ when she was one year old.

A. sing   B. cry   C. say   D. talk
(Key: C )

---

Table 2 Examples from the commonsense task

### 2.2.2 Test 2 Emotional inference test

To identify whether the participants and two versions of ChatGPT could recognize characters' emotional states, T2 adapted the emotional mental model story material (EMM) developed by

---
[4] Gaokao: College Entrance Examination in China

Gernsbacher et al.(1992). The 24 stories consist of 12 pairs of emotional states, including "Guilty-Proud, Bored-Curious, Sad-Joyful, Shy-Confident, Restless-Content, Afraid-Bold, Depressed-Happy, Disgusted-Admiring, Envious-Sympathetic, Callous-Caring, Desperate-Hopeful, and Angry-Grateful" (Gernsbacher et al., 1992, p. 95). In addition, these stories mainly describe the daily activities of adolescents, which may arouse our participants' interest and activate their life experiences to detect the characters' emotions. According to Gernsbacher et al., (1992), each story contains a unique emotion without any other implied emotions, avoiding disputing the answers. Table 3 illustrates the stories for emotional inference test in this study.

The stories have been widely adopted to test whether readers can infer specific emotions (e.g., Gygax et al. 2003, 2004). Gygax et al. (2003) found that participants may not always integrate exact words of emotions given by Gernsbacher et al. (1992), but they would recognize emotions consistent with the content of text. The study by Gygax et al. (2004) further indicated that readers had the capacity to infer specific emotions under certain conditions. Obviously, the material is feasible in this study.

The original material was comprised of two types of tasks(i.e., match/mismatch questions and filling stories) and was conducted using computer programs to test participants' reaction time (Gernsbacher et al., 1992). A bit different from Gernsbacher et al.'s (1992), the present study was an offline survey and did not include a measure of reaction time. During the test, participants were given enough thinking time to analyze the characters' emotions. Therefore, to avoid students from guessing the answers according to the matching or mismatching emotions designed by Gernsbacher et al. (1992), we changed the original task into the intuitive prompt, for example, "*At that moment, John felt_____*", in which each participant was required to fill in one word to best delineate the emotion of the person involved. We added this prompt at the end of each story as shown in Table 2. Moreover, some words that were new to the students were provided with Chinese translations to decrease the influence of the familiarity of vocabulary on their reading comprehension.

---

1. John, who always made good grades, had just transferred to a new school. He wished he had a hobby to occupy his time or something to keep him busy in the afternoons until he made more friends. After all, his new school was simply not very much of a challenge. And today was no different. As he walked home, he thought about another afternoon, just sitting around watching stupid reruns on TV.

   At that moment, John felt ________________________

9. For two days now, the snowstorm had confined Jackie to her small house. She paced from room to room. First, she went into the living room and picked up a book. She read two paragraphs and then put it down. Then she tried to find something on TV. After flipping the channels for fifteen minutes, she turned it off and wandered into the kitchen. Several times she opened the refrigerator, looked around, but then closed the door.
   At that moment, Jackie felt ______________

---

Table 3. Sampling stories adapted from Gernsbacher et al. (1992)

**2.2.3 Test 3 Causal inference test**

For the sake of testing the causal inferential ability of both participants, T3 selected a short story instead of giving a premise and hypothesis to find whether they can generate correct inferences by linking with the hidden clues expressed in the text. Therefore, a scary story full of suspense, named

*The Death Car* (Dagestani, n.d.), was adopted and modified, including adding a crucial inferential detail and some Chinese translations of words that participants hadn't learned and deleting the ending part for participants and chatbots to infer.

The condition we added is to describe the murderer's method of killing, with the original version being "*The man, John Downey, is a murderer who killed six people before he was captured two years ago.*" and the final version is "*The man, John Downey, is a murderer who killed six people by hanging (吊死) victims before he was captured two years ago.*" This is to help participants and GPTs to make more detailed inferences about the consequence of the story. Furthermore, we designed five questions with Questions 1, 3, and 4 evaluating participants' and GPTs' causal antecedent inferences, namely evaluating whether they can infer the causes of some phenomena. Questions 2 and 5 were tailored to examine whether they can reason the original ending.

---

I. Please make appropriate inferences according to the given text.

**Question (Q) 1. According to the story, what do you think might be the strange scratching noise?**
(Key: The killer was strangling George against the car roof and hanged him under the tree. )

**Q2. According to the story, what do you think is why George didn't come back last night?**
(Key：Because George was killed by the killer.)

**Q3. Why did the police come to surround the car?**
(Key: Maybe someone saw the dead body hanging above the car and called the police.)

**Q4. According to the story, what/who might cause the bump noise?**
A. the murderer    B. the branches    C. the raindrops    D. George's legs
(Key: D )

**Q5. Why did the police ask Marie to not look back?**
A. because she may see George's dead body hanging under the huge tree.
B. because she may see the murderer standing behind the car.
C. because the car was scratched badly by the branches.
D. because looking back is bad for a person who just woke up.
(Key: A)

Table 4: Questions and Answers of the Causal Inference Test

**2.3 Procedure**

With the objective of not interrupting the school's teaching schedule, we divided the whole data collection process into three stages and set them in the students' flexible class time when they did not have required classes. Before the three stages of the survey, we asked for participants' permission and then introduced our intentions and instruction to them. The first stage was the emotional inference test, during which participants were requested to finish the test within 45 minutes. The second stage was the causal inference test in which the students were given 15 minutes finish the task. The last stage contained 30 mins' answering time. To ensure fairness, all the participants were forbidden to communicate with each other during the whole test.

For the sake of comparison, both versions of ChatGPT were ordered to generate as many quantities of valid questionnaires as required, that is, 82 valid questionnaires for both emotional inferences and causal inferences, and 87 valid questionnaires for commonsense inferences. The excluded questionnaires were due to missing value. In addition, after comparing the initial causal inference results between two ChatGPTs and the students, we updated commands for the two ChatGPTs to improve their causal inference. The commands comprised four steps: (1) *sort out plotline of the story in terms of the following aspects: opening, build-up, climax, follow-up*; (2) *sort out key details that may influence the ending of the story*; (3) *who/what characters should be the focus in this story according to the plotline and details*; and (4) *based on the above analyses, revise your answers*.

We adopted accuracy to measure and compare their performances in the three types of inferential tasks, for accuracy is widely applied in previous contrastive computing studies (e.g., Mao et al. 2022, Guo et al. 2023).

## 3. Results

### 3.1 Commonsense inferences

Table 5 shows the accuracy of participants' and the two ChatGPT's performances in commonsense inferences: the students scored the highest in local cultures, with an average of 1.71 incorrect responses while ChatGPT Plus gained the highest mark in inferences of daily life experience, with only one incorrect response; the students made mistakes on average 4.85 questions in inferring daily life experience while ChatGPT offered incorrect responses to an average of 4.94 questions in local cultures and of 3.72 questions in daily life experience. ChatGPT Plus produced 4.34 incorrect responses to local culture inferences.

The present study also collected and summarized the explanations from the students, ChatGPT, and ChatGPT Plus to present why they made the inferences. For students, they turned to provide reasons like *"didn't know exact information"*, *"never eaten/seen/heard it"*, or *"have no idea"* when they didn't know the correct answers and then randomly made a choice. However, they seldom offered ambiguous reasons when they knew the answers. For example, they offered short but correct responses like *"passed away"* for Q1-Q4, *"it is sweet but not salty"* for Q7, *"not chicken eggs, but duck eggs"* for 11, and *"it gains its name because its shape resembles litchi"* for Q13.

By comparison, ChatGPT was inclined to make responses like *"the information cannot be verified"*, *"not enough information"* when it was not sure about the answers. What's worse, ChatGPT tended to fabricate facts to support its answers, like *"Liu Shaoqi passed away in 2012"*, *"Liu Shaoqi had ever taken a high-speed train"*, and *"Lu Xun did teach at Fuzhou University"* for 69 rounds in Q5, and *"it's indeed sweet and salty"* for questions concerning a traditional sweet porridge.

Additionally, ChatGPT Plus provided more accurate responses than ChatGPT, but not outdoing the students. for the first four questions, ChatGPT Plus did not generate any wrong reasons, i.e., *"He/She has passed away"* and respond only one incorrect reason in Q5, i.e., *"Lu Xun did teach at Fuzhou University"*. However, similar to ChatGPT, ChatGPT Plus could not identify wrong information. For example, it provided *"It is a unique traditional custom in Fuzhou"* for 87 rounds in 16, *"it's both sweet and salty"* for 65 times in Q7, *"it is a correct information"* for 74 rounds in Q10. Furthermore, ChatGPT also provided more accurate reasons for texts, such as *"Jesus is not included, so it's partly wrong"* and *"The cooking process is correct but it is not made from swallow"*. Notably, ChatGPT and ChatGPT Plus were discovered to make contradictory answers such as *"Liu Shaoqi had passed away in 1969"* with its answer *"T"*.

To conclude, the students showed relatively higher accuracy in inferring local characteristic cultures, whereas ChatGPT Plus showed definitely high accuracy in making inferences from daily life. When required to reason their answers, the students tended to offer reasons more correctly or stated their lack of experience while ChatGPT tended to fabricate facts or made contradictory answers when they could not make a judgment. By comparison ChatGPT Plus preferred to offer more accurate reasons for the texts available.

|  | Commonsense inference (average) | |
|---|---|---|
|  | Local cultures | Daily life experience |
| students | 86.83 | 67.66 |
| ChatGPT | 61.98 | 75.17 |
| ChatGPT Plus | 66.58 | 93.33 |

Table 5 General Performance on Emotional and Commonsense Inferences (Accuracy %)

### 3.2 Emotional inferences

Table 6 shows the results of responses to emotional inferences where the students, ChatGPT, or ChatGPT Plus made wrong inferences. Due to space limitation, here was presented the top-three-frequent words of each emotion. To unveil whether the responses were consistent with the emotions provided by Gernsbacher et al. (1992), we compared the main responses with the synonyms in the online dictionary of Merriam Webster Thesaurus (*Merriam-Webster: America's Most Trusted Dictionary*, n.d.) by the following standards: the selected synonyms were the best word choices for most users, based on machine learning and editorial review.

The comparison of performance between human readers and ChatGPTs revealed that the students failed to infer depressed, callous, caring, and gratitude. ChatGPT did not provide satisfied responses in shy, afraid, depressed, happy, callous, and admiration. In the meanwhile, none of the 82 rounds of ChatGPT Plus offered correct responses in shy, happy, callous, and anger. From the perspective of frequency, both versions of ChatGPT outpaced the students in all emotions, except for anger and gratitude.

Figure 1 shows the accuracy of emotional inference with regard to positive and negative emotions. The graph indicates that ChatGPT Plus outperformed both students and ChatGPT in inferring positive emotions, while students surpassed ChatGPT and ChatGPT Plus in negative emotions.

|  | senior-2 students | ChatGPT | ChatGPT Plus |
|---|---|---|---|
| shy | shy** (24), nervous (15), anxious (8) | nervous (61), anxious (25), insecure (9) | nervous (67), hesitant (23), anxious (21) |
| afraid | scared (25), afraid** (18), nervous (15) | cautious (45), anxious (40), apprehensive (12) | fearful* (65), anxious (58), nervous (14) |
| depressed | desperate (26), sad (12), helpless (9) | devastated (64), numb (14), defeated (8) | hopeless (56), devastated (26), depressed** (25) |
| happy | happy** (34), excited (33), proud (9) | ecstatic (78), elated (13), accomplished (11) | proud (55), elated (40), excited (15) |
| callous | angry (46), irritated (7), annoyed (6) | annoyed (67), indifferent (25), irritated (17) | indifferent (70), impatient (36), annoyed (26) |
| caring | happy (33), satisfied (25), relieved (9) | compassionate* (55), fulfilled (46), happy (10) | compassionate* (72), fulfilled (63), empathetic (5) |
| gratitude | moved (33), warm (18), happy (10) | grateful** (82), touched (15), supported (12) | grateful** (81), touched (39), loved (15) |

| | admiration | admired** (21), proud (12), surprised (6) | proud (59), impressed (19), inspired (11) | proud (77), impressed (59), admiring** (19) |
|---|---|---|---|---|
| | anger | angry** (41), sad (8), doubtful (7) | betrayed (82), hurt (30), angry** (5) | betrayed (82), hurt (41), humiliated (36) |

**: expected emotional words; *: synonyms of expected emotional words; numbers: number of correct answers or responding rounds

Table 6 Sampling responses to emotional inferences

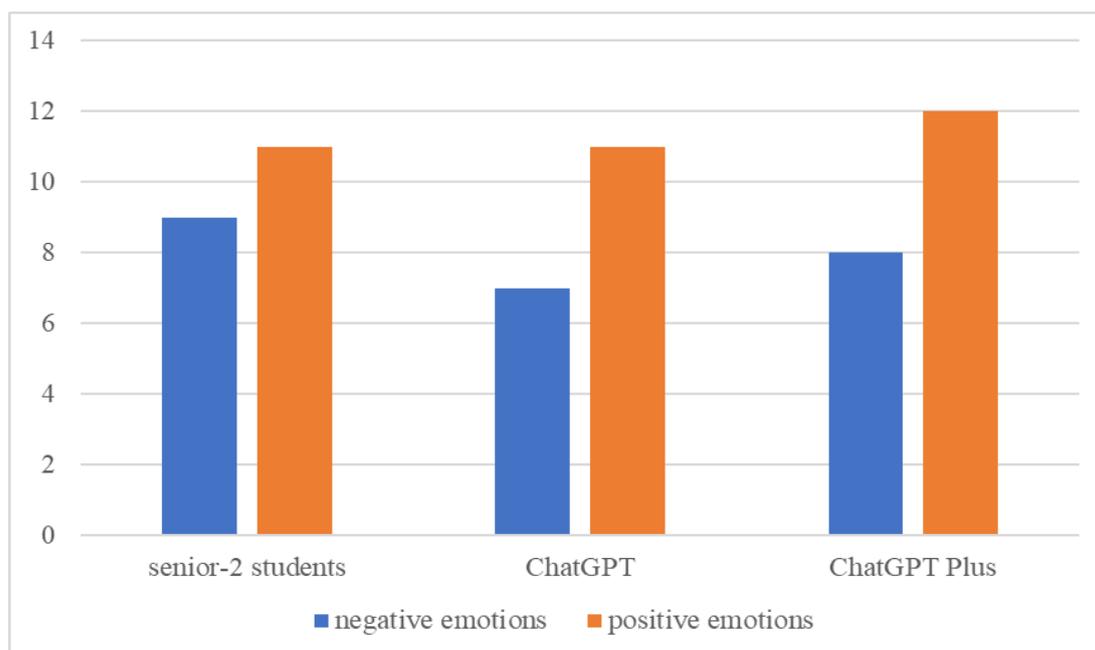

Figure 1 Accuracy of sentiment analysis

## 3.3 Causal inferences

As shown in Table 7, the students obtained the correct responses in Qs 1, 2, and 3, and the highest accuracy in Q4. ChatGPT failed at deducing Q1 and Q3, performed the worst in Q5, and offered inaccurate responses in Q4. ChatGPT Plus made inexact inferences in Q1 and Q3 but made correct prediction of Q2. Furthermore, it didn't output any correct responses in reasoning bump noises but performed the best in predicting the story ending, a bit superior to the students.

| | causal inferences ||||||
|---|---|---|---|---|---|
| | causal antecedents ||| causal consequences ||
| | Q1. cause of scratching noise | Q3. reason for the police's arrival | Q4. cause of bump noise | Q2. George's ending | Q5. Story ending |
| students | 1. by George's body/legs/hands/struggling (70)<br>2. murderer (7)<br>3. knife (4)<br>4. noise (1)<br>5. gunshot (1) | 1. George was killed/dead (56)<br>2. found George's dead body (18)<br>3. they caught the murderer/ there was a murderer/ search her husband/ save | 81.7 | 1. He died/was dead/killed by the murderer (82) | 96.34 |
| | | Marie/ Harrison called police (8) | | | |
| ChatGPT | 1. by the murderer (65) 2. by branches (16) | received report/information/tip/call about escaped murderer (82) | 2.44 | 1. encountered escaped murderer (44) 2. attacked or harmed by the murderer (21) 2. get lost (24) | 29.27 |
| ChatGPT Plus | 1. by the murderer to get into car (78) 2. by branches (23) 3. by the murderer (4) | 1. search for murderer (74) 2. found George's body or related to George's disappearance (34) 3. alerted by George | 0 | 1. harmed or killed by the murderer (36) 2. encountered John Downey (29) 3. became a victim of escaped murderer (15) 4. was caught by the escaped murderer (2) | 100% |

Table 7 Performance of Students and ChatGPTs on the causal inferences

### 3.4 Comparison between two ChatGPTs' causal inference

The results showed that ChatGPT Plus offered accurate inferences as the students did, but ChatGPT kept unchanged in the task. As the command was updated elaborately, ChatGPT Plus corrected its responses to the five questions. *"perhaps the sound of George's body or the murderer's interaction"* for Q1, *"because he encountered the murderer...became a victim"* for Q2, *"searching for the escaped murderer and found...crime scene"* for Q3, *"George's legs"* for Q4, *"she may see George's dead body"* for Q5. These responses indicated that ChatGPT Plus amended its' mistakes in making antecedents and consequences. On the contrary, ChatGPT generated responses consistent with previous answers, i. e. *"the branches"* for scratching noise and bump noise, *"he might have been captured or harmed by the murderer"* for Q2, *"heard the knocking sounds and saw the parked car"* for Q3, and *"she may see the murder standing behind the car"* for Q5.

To better understand why the two chatbots perform differently, we requested the two versions to provide their reasons for making inferences. Evidently, ChatGPT Plus inferred the answers from key details and wove the details into a complete ending while ChatGPT just made inferences from single plots but without a holistic view of the storyline. More specifically, ChatGPT Plus inferred the plots by summarizing seven key details and cues, namely *"the scratching noise & continuous knocking"*, *"George's Absence"*, *"Police arrival and warning"*, *"John Downey's instruction"*, *"Marie's central role"*, *"Car's location and condition"*, and *"George's instructions to Marie"* to make up the whole of suspense story. By contrast, ChatGPT analyzed the plot based on explicit plots but not stringing them together. To be precise, ChatGPT offered *"car ... under a huge tree"*, *"he left Marie... and didn't return...the story...creates a sense of potential danger in the environment"*, *"news about the escaped murderer...police issuing a warning"*, *"car parked under a huge tree"*, and *"potential danger...aligns with ... threat of the escaped murderer"* for the five questions respectively, suggesting that ChatGPT did not change its inferences according to the updated command.

## 4. Discussion

The present study compared the performances on three different inferences (commonsense, emotional, and causal inferences) by China's senior school students, ChatGPT and ChatGPT Plus by virtue of English text reading comprehension, and then analyzed the results according to the accuracy in their responses so as to reveal the advantages and disadvantages by both human readers and the chatbots in processing texts. The whole study consisted of three tests, respectively regarding commonsense inference, emotional inferences, and causal inferences.

Results revealed that ChatGPT Plus gained the best performance in making daily-life inferences and emotional inferences, whereas ChatGPT performed worst in the three inferences. In addition, the students showed the best performances in commonsense inference concerning local culture and causal inferences, but did worse in commonsense inference regarding daily life and emotional inferences. These results unveil the inference capacity of two ChatGPT versions and the students, hence acting as the answer to the first question.

Furthermore, our data presented the advantages and disadvantages of ChatGPTs and the students in making inferences, therefore answering the second research question. Specifically, ChatGPT was weak at making inferences requiring specific world knowledge, concerning subjective judgements, or related to logical organization while ChatGPT Plus was skillful at dealing with deductive inferences and questions that requested large lexical storage and language capacity but faltered in making inferences out of its knowledge domain. By contrast, the students were adept at inferences within their knowledge and making logical reasonings, yet they were poor at making inferences involving difficult grammar and vocabulary analyses.

Finally, the causal inference comparison of the two ChatGPTs under the four rounds of updated commands unlocks the point that ChatGPT Plus could improve its responses while ChatGPT kept unchanged, answering the third question we posed. These findings converge to suggest that ChatGPTs and the students were complementary in handling inferences in narratives. The following is to elaborate on what may account for the findings in this study.

### 4.1 Human readers superior to ChatGPTs in commonsense inferences

As shown above, the accuracy confirmed that the students were superior to both ChatGPT and ChatGPT Plus in inferring local cultures but lagged behind the two chatbots in detecting daily-life inferences. This result is consistent with Qin et al. (2023) that ChatGPT did not always offer better performance gains in commonsense inference tasks. The major reasons are concluded as follows.

First of all, the students analyzed texts' implicit meanings whereas ChatGPTs comprehended texts based on literal meaning. When reading a text, the students tended to understand sentences by linking both words and information together to achieve coherence (Kendeou et al., 2014) and by activating complicated cognitive processes and prior knowledge (Perfetti & Stafura, 2014). That is, they did not focus only on literal information but combined all resources to make correct judgments, resulting in their high accuracy in identifying the incorrect parts in the texts. By contrast, if one sentence does not contain obvious mistakes like "*the sun is square*", or grammatical errors, the two ChatGPTs would presuppose the correctness of the sentence, conforming to Kojima et al. (2023) that ChatGPT's answers may include the mistakes that only humans can identify. For example, in the fourth round generation under the same command, ChatGPT provided the response, "*General Liu Shaoqi passed away in Beijing in 2012, but his death was not caused by taking the high-speed train to Changsha.*", demonstrating that ChatGPT made the judgement in light of text's literal meaning while unknowing the knowledge as common sense (concealed in the text) to human readers. Similarly, in the 12th, 65th, and 78th rounds of output relating to Q1, ChatGPT explained that "*The news is true because it is reported by Xinhua News Agency.*" This situation concurs with OpenAI's tests on separating fact from incorrect responses by GPT-3.5, namely, GPT-3.5 did not perform well in identifying false information (OpenAI, 2023). Why? First, GPT-3.5 generally lacks world knowledge. Second, GPT-3.5 would accept users' false information and lack the ability to judge facts from false information (OpenAI, 2023).

Next, while the two ChatGPTs lack the capacity to distinguish right from wrong, the students had the ability to identify true information. Qs 7 and 8 were specially designed containing contradictory information, namely, "Aojiu congee is sweet and salty" and "Aojiu congee is sweet". This design was to reveal whether the students, ChatGPT, and ChatGPT Plus could recognize the true statement. As expected, the students provided 78 responses of "*it is sweet*" for both questions, indicating that most students can made the correct judgement. On the basis of their commonsense (for example Aojiu congee is similar to Laba porridge, an important sweet dessert served during the festival.) Unexpectedly, However, ChatGPT and ChatGPT Plus regarded Q7 ("sweet and salty") as correct information and hence responded to Q8 in terms of the Q7. More specifically, ChatGPT generated *"It's indeed sweet and salty"* for 41 rounds and 13 rounds in Qs 7 and 8 respectively, and ChatGPT Plus generated *"It's both sweet and salty"* for 65 times and 62 times in Qs 7 and 8 respectively. These instances suggest that ChatGPT and ChatGPT Plus are weak at making correct judgement of some information provided, which may be harmful when they input false information with ill intentions. This was also noticed by OpenAI (2023, p. 10) that GPT-4 can sometimes "be overly gullible in accepting obviously false statements from a user", let alone its inferior version GPT-3.5.

In addition, the students preferred definite responses while ChatGPT Plus may offer ambiguous answers. When the students spot wrong information in the statement, they were accustomed to deciding the statement as a false one. Meanwhile, the students were loyal to the instruction, i. e. make True or False judgment. On this account, they were impossible to provide neutral responses. Yet different from the students, ChatGPT Plus showed a more unsteady fashion when explaining its answers in detail, like *"partly right/wrong"*. The rigorous attitude is traceable. According to OpenAI (2023), the OpenAI research team had significantly improved GPT-4 from the perspective of reducing hallucinations and common sayings. This may contribute to the fact that ChatGPT Plus was inclined to provide more accurate responses.

But ChatGPT Plus demonstrated higher proficiency than ChatGPT in inferring commonsense. Although ChatGPT Plus did not excel ChatGPT by a large margin in the accuracy of inferring the local cultures, it did display its progress in reasoning from texts. This aligns with the previous finding that GPT-4 outperformed ChatGPT in exhibiting common sense (Bubeck et al., 2023), similarly owing to its much larger pre-training datasets.

The explanations of the given statements further revealed the features of responses by the students and ChatGPTs. To begin with, the students were honest: when they did not know the information, they tended to give responses like *"have no idea" or "never heard before"*, as if more uncertain and subjective. By contrast, GPTs would generate definite responses and look more objective (at least on the surface), such as *"the information cannot be verified" or "no enough information"*. Secondly, the students' explanations were inclined to be short and simple, whereas the GPTs' were long and deliberate, consistent with Guo et al. (2023). For example, in the Qs 1-4, the students offered "*He/She has passed away*" whereas ChatGPT Plus provided "*Qian Xuesen, a famous scientist in the field of rocket and space technology, passed away in 2009. Thus, he cannot appear on the program.*", and ChatGPT generated "*Qian Xuesen, also known as Hsue-Shen Tsien, was a prominent Chinese scientist and engineer who passed away on October 31st, 2009. He cannot record a program in 2023.*" This discrepancy may result from the different strategies adopted in answering questions: the students were accustomed to pointing out the illogical fallacy directly while GPTs decoding the statement step by step according to the given knowledge in the datasets (Guo et al., 2023; X. Wang et al., 2023).

Additionally, when reasoning the commonsense concerning local cultures, ChatGPT may choose to concoct facts. This phenomenon was also noticed by Guo et al. (2023) that ChatGPT tended to fabricate facts when professional or specific knowledge from a particular field was needed to answer a question and by OpenAI (2023) that ChatGPT would hallucinate facts. Accordingly, it is reasonable to argue that the local cultures are out of the range of ChatGPT's training datasets, leading to the result that the chatbot can either provide a false response to our request or make an unrealistic statement.

Another strange finding is that the two ChatGPT versions were likely to provide contradictory answers, though not on a large scale. For example, in the 67$^{th}$ round of Q2, ChatGPT offered "*T. Qian Xuesen passed away on July 31$^{st}$, 2010, therefore this statement is correct.*", a response contradictory to reality; and ChatGPT Plus provided a similar explanation "*T. The dish Lichee Pork (荔枝肉) gets its name from its litchi-like appearance when cooked, not because of litchi is added to the recipe*" in the 64$^{th}$ round of Q13, which was conspicuously against the matter of fact. This might be caused by the scale and range of its pre-training datasets, or the misprocessing of context. This finding corresponds with the analyses of Guo et al. (2023, p. 6) that ChatGPT "refuses to answer the question out of its knowledge". By contrast, ChatGPT Plus due to its greatly extended pre-training dataset became able to generate more inferential responses in accord with what things are, adding certainty in reasoning judgment.

With regard to the daily life inferences, ChatGPT Plus surpassed both the students and ChatGPT remarkably, fully in line with Zhu et al. (2023). According to the report by OpenAI (2023), GPT-4 gained great performances in various academic benchmarks specially designed for humans. To our expectation, the students' poorer performance results from the fact they are still at the high school level and hence do not possess a large vocabulary and a relatively premium capacity to analyze the context provided. And this in turn led to the lowest scores in multiple-choice test for both text comprehension and lexical meaning cognition.

**4.2 ChatGPT Plus outweighing human readers in emotional inferences**

Our statistical data revealed that ChatGPT Plus outperformed the students in inferring specific emotions but ChatGPT fell behind them on the whole. Specifically, the students did not do better than ChatGPT Plus but outdid ChatGPT in terms of accuracy. In addition, the students were inferior to the two ChatGPTs in frequency of inferring emotions where the three parties all generated correct responses. This suggests that ChatGPT Plus outweighed human readers with intermediate English proficiency in making emotional inferences, which was also present in Elyoseph et al. (2023) and Kocon et al. (2023). Furthermore, similar to the research by Gao et al. (2023), the students detected negative emotions more accurately. To the opposite, the two ChatGPTs judged the positive emotions more correctly than students, quite similar to Kabir et al. (2023) that ChatGPT significantly generated less negative emotions than human beings when answering questions.

The reason why ChatGPT Plus outweighed the students may be that the students possessed inadequate vocabulary and reading comprehension ability. Vocabulary and text-reading ability have been recognized to influence students' inference-making ability (Barnes et al., 2015; Li & Kirby, 2014; Oslund et al., 2016; Prior et al., 2014). Although the students participating in the present study had learned nearly 5000 English words and various reading texts, their English proficiency lagged much behind the two ChatGPTs, according to the statistics by OpenAI (2023). In addition, the two ChatGPTs have been examined by diverse datasets of emotions and sentiments (Bang et al., 2023; Guo et al., 2023; Qin et al., 2023), consequently extracting various emotion types more skillfully. That may explain why the two ChatGPTs understood the stories better than the students. Additionally, their responses further illustrated the different performances in judging emotions concealed in the texts:

Compared with the two ChatGPTs, the students could sense the nuances of emotions implied in the context. The results showed that the students produced much more emotional words than the two ChatGPTs (25.42 by students, 7.45 by ChatGPT, and 7.33 by ChatGPT Plus), manifesting that human readers are sentimental creatures and hence could sense more delicate and subtle feelings hidden in the contexts compared with AI chatbots (Carlbring et al., 2023). For example, in story 17, the 82 students showed 38 feelings, such as "*shocked*", "*amazed*", "*incredible*", and "*uncomfortable*", which may be their first impression when answering the phone. This result is in line with the research of Gygax et al. (2003) that their participants inferred different emotions with an average of 26 items. By contrast, the two ChatGPTs output fewer words regarding emotional inferences. Basically, no change was observed in their responses we ran the same tasks in ChatGPT and ChatGPT Plus several times. For example, ChatGPT provided only two words (*bored and lonely*) for the boredom detection after 82 rounds and ChatGPT generated four words (*betrayed, hurt, angry, and humiliated*) for anger inference.

The two ChatGPTs relative to the students would offer broader and more formal words to exaggerate characters' emotions in texts. For example, when inferring *happy*, ChatGPT offered "*ecstatic*" for 78 times and "*elated*" for 13 times, which implied a much higher degree of happiness. As for *anger*, the two versions inferred "*betrayed*", deviating and overstating the character's angry mood. Cases as such contained "*assertive*" for confident, "*proud*" and "*elated*" for happy by ChatGPT Plus, and "*empowered*" for bold, "*devastated*" for depressed, and "*humiliated*" for anger by ChatGPT. This situation may be because "a machine does not yet possess human-like empathy or emotions, and hence has difficulty understanding the nuances of human language." (Carlbring et al., 2023, p. 1).

**4.3 Human readers' better performance than ChatGPTs' in causal inferences**

Test 3 adapted a suspense story by deleting the ending for consequence inference and tailoring questions for antecedent inferences based on the text. The results showed that the students outweighed ChatGPT and ChatGPT Plus in detecting antecedents while ChatGPT Plus won the students by a slight margin in inferring story ending. Therefore, these results suggested that human readers outperformed the two ChatGPTs in making causal inferences. The finding provided evidence for Liu et al. (2023) that GPT-4 did not master all types of logical reasoning, and its performance in logical reasoning of natural language was not as strong as it was in multichoice reading comprehension. However, this finding is contrary to Bang et al. (2023) that ChatGPT was excellent in making causal inferences, which may result from the large pre-trained datasets, namely the datasets they used may have been encoded in ChatGPT.

In Test 3, the students surpassed two ChatGPTs in analyzing the antecedents and consequences. The students centered their focus on the couple and murderer by connecting events from the whole text, which coincides with previous opinions that causal-inference making would help link successive events together (Mason & Just, 2004). According to their responses, the students process the story based strictly on the principle that antecedents happen before consequences (Van Den Broek, 1990). As a result, the students inferred the antecedents of scratching sounds and bumping sounds based on the murderer's killing method and the environment: a huge tree, thus making the inference that the sounds were caused by George's body when he was strangled by the escaped murderer. Furthermore, they inferred the consequences according to previous episodes: the escaped murderer, George's missing, and the police's arrival. Without these antecedents, the consequence would not happen. Along this storyline, the students deducted their answers by virtue of the detailed description like *"a murderer...by hanging victims"*, *"under a huge tree"*, *"coming from the roof"*, and *"the knocking had never stopped"* *"Why had he not come for her?"* *"look straight ahead...don't look back"*. To conclude, the students drew the inferences from the following aspects: (1) announcement about the escaped murderer, indicating murder would happen in the near future, (2) murderer's killing method, car's location, continuous noises coming from the roof, George's instruction, suggesting murder related with hanging occurred, (3) police arrival and instruction, George's missing, implying the case should be correlated with Marie. This inference-making logic was conformed to Alonzo et al. (2009) that in order to make correct inferences, readers should recognize and comprehend the inner connections among objects, events, or characters implied in the text.

In contrast with the students, ChatGPT did not detect both antecedents and consequences correctly. This was foreseeable since ChatGPT was recognized as not fully understanding the meaning behind words and lacking analytical thinking ability (Farrokhnia et al., 2023). ChatGPT's responses uncovered that the chatbot analyzed the antecedents based on the plot in the immediately prior context but not on the whole story. For example, in Qs 1 and 4 about the causes of noises, ChatGPT extracted information about trees and the murderer but did not connect them with the subsequent plot, such as George's instructions and disappearance, the murderer's killing method, and the police's arrival. Similarly, in Qs 3 and 5, although ChatGPT did connect the murderer to police to infer that a murder would have happened ChatGPT foundered on correlating the murder to George's disappearance. In summary, ChatGPT made the inferences on the basis of the details: *"a man escaped" "a murderer"*, *"under a huge tree"*, and *"Marie locked the door"*, with which the storyline would become a police-catching-murderer background instead of a story centered on the couple, ignoring the role of story's protagonists.

Similar to ChatGPT, ChatGPT Plus did not infer the antecedents in the contexts but performed well in detecting the story ending. This finding is in line with Liu et al. (2023) that GPT-4 did not always perform well in inferring natural language. The answers concerning sources of noises (Qs 1 and 4) showed that the chatbot drew the inferences concerning antecedents from the prior context, namely the murderer escaped and the environment but without considering George's subsequent fate. However, one of the answers of Q3, *"found George's body or related to George's disappearance"*, displayed that it connected the escaped murderer and police with George's disappearance, suggesting that compared with ChatGPT, ChatGPT Plus had stronger logical ability in inferring causal antecedents. As for the story ending, ChatGPT Plus successfully deducted the correct storyline connecting details of *"a man escaped"*, *"a murderer"*, *"under a huge tree"*, *"Why had he not come for her?"*, *"Several policemen leapt out"*, and *"look straight ahead...don't look back"*. These clues helped ChatGPT Plus infer the ending by integrating the details into the whole text and analyzing the context more naturally. In addition, according to the testing data, ChatGPT Plus performed better in causal consequence than in causal antecedent, for ChatGPT is better at making conclusions than at generalizing rules after specifically observing the given information (Bang et al., 2023).

Overall, human readers had a higher capacity to extract crucial information from a text to trace both hidden causes and results, suggesting their better logical ability in comprehending narrative than ChatGPT and ChatGPT Plus. Moreover, when making inferences, human readers prefer to extract clues from the whole text and consider the context whilst ChatGPT and ChatGPT Plus lack the ability to sort out significant details relevant to the storyline and integrate them into an intact story. In addition, when analyzing the antecedents, both ChatGPTs seem indifferent to the influence of subsequent plots, resulting in incorrect inferences.

**4.3 ChatGPT Plus Outdoing ChatGPT in inferences under updated commands**

Our test revealed another finding that ChatGPT Plus started to make correct inferences under more elaborated commands, but ChatGPT did not demonstrate favorable inferential logic. This finding is consistent with Liu et al. (2023) that "ChatGPT is not good at following NLI (Natural language inference) task instructions" while ChatGPT Plus was able to correct answers when its command was updated.

In the test, ChatGPT Plus corrected its logical mistakes and made reasonable causal inferences. Results showed that ChatGPT Plus was able to generate not only causal antecedents but also causal consequences accurately by connecting details and successive events in the story. The improvement could be caused by its updating processes: GPT-4 was tested on various human examinations and benchmarks to enhance its natural language generation and comprehension (OpenAI, 2023).

By contrast, ChatGPT still focused on the literal responses of the story. For example, when it comes to scratching noise, ChatGPT generated "*noise could be a natural occurrence due to the branches coming into contact with the car's roof*", indicating that it made the antecedent inferences based on the noise itself without considering the contextual clues and terrific atmosphere. As for the bumping noise, ChatGPT responses that "*branches in the wind could cause such noises, aligning with the atmospheric tension of the story*" demonstrating that it made the causal consequence according to the continuous noise and Marie's hiding in the car without connecting the antecedents. The reason why ChatGPT could not gain self-improvement may lie in its lack of deep understanding and high-order thinking skills (Farrokhnia et al., 2023), leading to responses off-topic (Gupta et al., 2023).

In summary, while ChatGPT acts as a "stupid" reader who is an expert in reasoning with literal text but fails to draw deep inferences by using all the cues in the whole story, ChatGPT Plus could be led to make reasonable inferences by more precise commands or instructions.

## 5. Conclusion

This research undertook a comparison of Chinese high school students, ChatGPT, and ChatGPT Plus in their abilities to draw inferences from reading English narrative texts in the perspective of three dimensions (commonsense, emotional, and causal interpretations).

While ChatGPT and ChatGPT Plus exhibited high proficiency in making commonsense inferences concerning everyday experiences, they faltered when processing knowledge beyond their training datasets and occasionally misinterpreted texts, or even fabricated facts. In causal inference, particularly in detecting causal antecedents and consequences, human readers represented by ESL Chinese students clearly outpaced both AI models. However, when offered elaborately refined commands, ChatGPT Plus demonstrated good causal inference ability while ChatGPT did not. Emotionally, the two ChatGPT versions demonstrated better capacity to detect emotions from given texts in light of frequency. By contrast, provided that all emotions were calculated accurately, ChatGPT Plus and human readers displayed equal capabilities, both surpassing ChatGPT. Evidently,

ChatGPTs are complementary to human readers in reasoning with English text reading comprehension.

In conclusion, this study highlights the power of AI models like ChatGPT and ChatGPT Plus in textual inferences within their training datasets. Meanwhile, it also illuminates their limitations in more nuanced reading comprehension tasks and underscores human readers' superiority in this regard. Such insights pave the way for future refinements in subsequent AI advancements.